\documentclass[10pt,twocolumn,letterpaper]{article}
\pdfoutput=1
\usepackage{cvpr}
\usepackage{times}
\usepackage{epsfig}
\usepackage{graphicx}
\usepackage{amsmath}
\usepackage{amssymb}
\usepackage{comment}

\usepackage[T1]{fontenc}% for special charactors
\usepackage{flushend}
\usepackage{fixltx2e}
\usepackage{url}
\usepackage{multirow}
\usepackage{verbatim}   % useful for program listings
\usepackage{amsmath}    % need for subequations
\usepackage{amsfonts}
\usepackage{subfig}

\DeclareMathOperator*{\argmin}{arg\,min}

\newcommand{\savespace}{\vspace{-10pt}}

\cvprfinalcopy % *** Uncomment this line for the final submission

\begin{document}

%%%%%%%%% TITLE
\title{Transformed Residual Quantization for Approximate Nearest Neighbor Search}

\author{Jiangbo Yuan\\
Florida State University\\
{\tt\small jyuan@cs.fsu.edu}
\and
Xiuwen Liu\\
Florida State University\\
{\tt\small liux@cs.fsu.edu}
}

\maketitle

\begin{abstract} 
The success of product quantization (PQ) for fast nearest neighbor search depends on the exponentially reduced complexities of both storage and computation with respect to the codebook size. Recent efforts have been focused on employing sophisticated optimization strategies, or seeking more effective models. Residual quantization (RQ) is such an alternative that holds the same property as PQ in terms of the aforementioned complexities. In addition to being a direct replacement of PQ, hybrids of PQ and RQ can yield more gains for approximate nearest neighbor search. This motivated us to propose a novel approach to optimizing RQ and the related hybrid models. With an observation of the general randomness increase in a residual space, we propose a new strategy that jointly learns a local transformation per residual cluster with an ultimate goal to reduce overall quantization errors. We have shown that our approach can achieve significantly better accuracy on nearest neighbor search than both the original and the optimized PQ on several very large scale benchmarks.
\end{abstract}

\section{Introduction}
\label{setI}

Nearest neighbor search in very large databases is becoming increasingly important in machine learning, computer vision, pattern recognition, and multimedia retrieval along with many applications in document, image, audio, and video retrievals~\cite{tLiu2004, Weiss2008, Lowe2004, Datar2004, Andoni2006, Nister2006, Philbin2007, Boiman2008, Jegou2008, JegouDYGG12, HeKC2012}. However, it becomes difficult to efficiently store and search huge collections when the dataset size gets larger and larger (e.g. millions or even billions). The idea of mapping real-valued vectors to compact codes \cite{Torralba08smallcodes, Jegou2010, GongS2011} provides a very attactive way to solve this problem. There have been many recent methods on developing hashing methods for compressing data to compact binary strings \cite{Torralba08smallcodes, GongS2011, Andoni2006, Kulis2009, Charikar02,SSH:cvpr2010,cvpr12:KSH}, or vector quantization based methods \cite{Jegou2010, GeHKS2013, NorouziF2013} that compress data to compact codes.

Vector quantization (VQ) was actively studied for source coding/signal compression under real-time constraints dating back to decades ago~\cite{Gersho-VQ-Book}. Recently, the problem of how to apply VQ techniques to efficient approximate nearest neighbor (ANN) search has attracted a lot of attension~\cite{Jegou2010, chen2010, YuanL12, BabenkoL2012, GeHKS2013, NorouziF2013, ZhangDW2014, BabenkoL2014}. While many structured VQ models can be found~\cite{Gersho-VQ-Book, JainMF1999, BarnesRN1996}, in this paper, we restrict our attention to product quantization (PQ) and residual quantization (RQ), which both have been successfully applied to fast nearest neighbor search \cite{Jegou2010,chen2010}.

\subsection{Related Works}
Product quantization works by grouping the feature dimensions into groups, and performs quantization to each feature group. In particular, it performs a $k$-means clustering to each group to obtain sub-codebooks, and the global quantization codebook is generated by the Cartesian products of all the small sub-codebooks. In this way, it can generate a huge number of landmark points in the space, which guarantees low quantization error; it has achieved state of the art performance on approximate nearest neighbor search~\cite{Jegou2010}, and can also provide a compact representation to the vectors. Inspired by the success of PQ, some latest works have extended PQ to a more general model by finding an optimized space-decomposition to minimize its overall distortion~\cite{GeHKS2013, NorouziF2013}. A very recent work~\cite{KalantidisA2014} has deployed this optimized PQ within residual clusters. While it maximizes the strength of locality, it also uses extra space for multiple transformations as well as PQ codebooks.

\begin{figure*}[th]
\centering
\includegraphics[trim=2cm 12cm 4cm 1cm,clip=true, width=.95\linewidth]{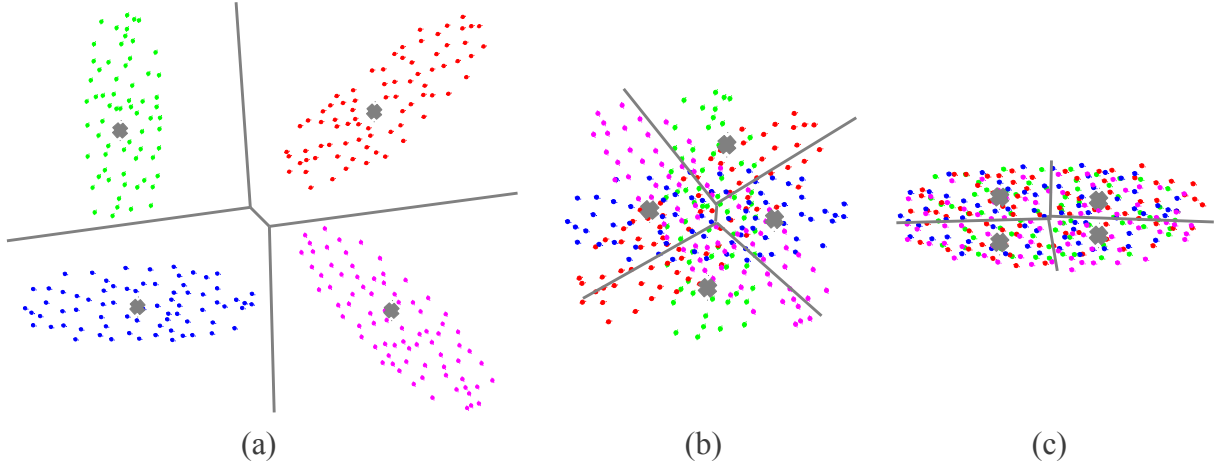}
\caption{A toy example of two-level RQ and TRQ. While the first-level vector quantizer (a) is identical to the two models, the residual space in the ordinary RQ model (b) is much noisy than the TRQ model (c), in which the residual clusters are aligned by our novel method.}
\label{figure:ORQ-example}
\end{figure*}

Different from PQ, RQ works by performing quantization to the whole feature space, and then recursively apply VQ models to the residuals of the previous quantization level, which is a stacked quantization model. In particular, it performs a $k$-means clustering to the original feature vectors, and construct $k$ clusters. For points in each cluster, it computes the residuals between points and the cluster centers. In the next level, it aggregates all the residual vectors for all points, and performs another clustering to these residual vectors. This process is recursively applied (stacked) for several levels. In this way, RQ produces sequential-product codebooks. A comprehensive survey of earlier RQ models can be found in Barnes et al.~\cite{BarnesRN1996}. Recent works have shown the effectiveness of RQ to perform both indexing~\cite{YuanL12} and data compression~\cite{chen2010} tasks in ANN search problems. 

However, it has been observed that the effectiveness of RQ might be limited to a very small number of levels~\cite{Gersho-VQ-Book}, and the randomness of residual vectors increases very quickly when we stack more and more RQ layers. This results in increasingly noisy residual vectors of each level, and effective quantization at higher RQ levels becomes more difficult. In other words, the layer-wise learning in RQ models provides suboptimal sub-codebooks in that each sub-codebook is learned without consideration to subsequent layers. Thence several global optimization approaches have been proposed by jointly learning sub-codebooks over all layers~\cite{ChanGG1992, ZhangDW2014, BabenkoL2014}. Different from seeking out joint optimization solutions, we here look for specific transformations to each of the residual clusters to make the resulting vectors more aligned, in order to directly address the problem of cluster misalginment and noise increase over each level after the first and improve overall quantization accordingly.

\subsection{Contributions}
To this end, we propose a novel approach to optimizing the RQ model that is motivated by the noise and shape in the residual space, as shown in Figure \ref{figure:ORQ-example}. We find that the residual vectors for each cluster have significantly different directions, which make quantization at the next level much harder. A natural idea is to align these residual vectors for each cluster, which can potentially reduce quantization error in the next level. Thus, we propose to learn one rotation matrix for each residual cluster, and use them to align the residual vectors in each cluster, so as to reduce the global quantization error. Then we alternate between learning rotation matrices and the residual quantization to minimize its distortion, which is mainly inspired by ``Iterative Quantization (ITQ)'' \cite{GongS2011}; very recently ITQ has been successfully applied to the PQ models~\cite{GeHKS2013, NorouziF2013, KalantidisA2014}. Different from ITQ and the optimized PQ (OPQ), which learn a global rotation, both ours and the locally optimized PQ (LOPQ)~\cite{KalantidisA2014} learn one projection matrix per residual cluster. In stead of independent learning in LOPQ, however, all local transformations in our method are learned associatively with a shared codebook. In this way, we successfully reduce the memory overheads about codebook usage in LOPQ. This offers a great flexibility to use reasonable larger codebooks in potential.

We have found that by doing this iterative alignment and quantization, we can achieve significantly smaller quantization error than both vanilla RQ and OPQ methods. In addition, we also propose a hybrid ANN search method which is based on the proposed TRQ and PQ. Experimental results on several large-scale datasets have clearly demonstrated the effectiveness of our proposed methods, in particular have shown that the extra transformations only introduced very limited computation overheads when integrated with advanced indexing structures, e.g. the inverted mutli-index~\cite{BabenkoL2012}. Therefore, our method is able to achieve very significant gain over other state of the art methods in terms of trade-off among memory usage, search quality and speed.

\section{Background and Formulations}
\label{sectR}

In this section, we review related background on two types of structured vector quantization, i.e. product quantization and residual quantization. To accurately quantize large number of points in a high dimensional space, we need a large number of landmark points (or centroids). For example, if we use the $k$-means clustering method to find such quantizers, when the number of cluster centers $k$ becomes huge (e.g. millions or even more), performing $k$-means becomes prohibitively expensive. Structured vector quantization make special assumptions about the data distribution, and try to explore such structure to generate large number of landmark points to increase the quantization accurateness \cite{Gersho-VQ-Book}. PQ and RQ are two structured VQ families with different assumptions about data distribution. We here present discussions on structural codebook constructions and objective function formulations of the related VQ models following a brief introduction of the unstructured VQ.

Vector quantization (VQ a.k.a.~$k$-means) without any structure constraints is probably one of the most widely used vector quantization method. Given a dataset $X = \{ \boldsymbol x_{j}: \boldsymbol x_{j} \in \mathbb{R}^{D}, j = 1,...,N\}$, VQ is a mapping:
\begin{equation}
\label{eq:vq_mapping}
q(\boldsymbol x_{j})= \boldsymbol c_{i}, \boldsymbol c_{i} \in C
\end{equation}
where $q$ is a quantizer and $\boldsymbol c_{i}, i = 1,2,...,k$ is a \textit{centroid} or a \textit{center} from the codebook $C$. According to Lloyd's optimality conditions, an optimal quantizer satisfies the nearest neighbor condition: $\displaystyle q(\boldsymbol x_{j}) = \argmin_{\boldsymbol c_{i} \in C} d(\boldsymbol x_{j}, \boldsymbol c_{i}).$ Here $d(., .)$ is the distance between two vectors and Euclidean distance is used in this paper. For each centroid, a set of data points will be assigned to it, and forms a cluster. According to the second optimal condition, a centroid is computed as: $\boldsymbol c_{i} = E(\boldsymbol x_{j} | \boldsymbol x_{j} \in V_{i}).$ We estimate the codebook centers to minimize the objective function: the mean squared error (MSE)
\begin{equation}
\label{eq:vq_mse}
\text{MSE}(q)= 1/N \sum_{j=1}^{N} \left \| \boldsymbol x_{j}-q(\boldsymbol x_{j}) \right \|_{2}^{2}.
\end{equation}

While the globally optimal solution of the above problem is NP-hard, it can be solved by heuristic alternatives. The best known approach is the $k$-means algorithm~\cite{Gersho-VQ-Book}, in which the above two conditions are optimized alternatively.

\subsection{Product Quantization}

Product quantization assumes that certain groups of features are independent with each other, and explores this assumption to generate a large number of landmark points by grouping the feature space into $m$ groups (each group is a subspace). By quantizing each of the subspaces separately using sub-quantizer $q_i(\cdot)$, it produces an implicit codebook as a Cartesian product of $m$ small sub-codebooks, $C = C_{1} \times C_{2} \ldots \times C_{m}$. In this case, codebook $C$ can provide an exponentially large number of cluster centers while retains a linear size of storage. Given a data point $\boldsymbol x_{j}$, it estimates the globally nearest center by concatenating all its nearest sub-centers from the sub-codebooks as $q(\boldsymbol x_j) = \boldsymbol c_{i1} | \ldots | \boldsymbol c_{im} $.
The MSE of this product quantizer can be estimated as:
\begin{equation}
\label{eq:pq_mse}
\text{MSE}(q) = \sum_{i=1}^{m} \text{MSE}(q_{i}).
\end{equation}

The degradation of PQ performance can be severe if there are substantial statistical dependences among the feature groups~\cite{Gersho-VQ-Book}; Recent works have shown that such dependences could be reduced, to some extent, by more careful space-decomposition~\cite{GeHKS2013, NorouziF2013}. Their works extended the idea from iterative quantization~\cite{GongS2011} to the PQ scheme that learn an rotation to transform the data to reduce the dependences between feature groups. For example, the work from \cite{GeHKS2013} jointly seeks a whole space rotation $T$ and PQ codebooks $C$ by minimizing
\begin{equation}
\label{eq:opq_mse}
\text{MSE}(q)= 1/N \sum_{j=1}^{N} \left \| T\boldsymbol x_{j}-q(T\boldsymbol x_{j}) \right \|_{2}^{2},
\end{equation}
\savespace
\[
s.t.~~~~q(T \boldsymbol x_{j}) = T(\boldsymbol c_{i_{1}} | \boldsymbol c_{i_{2}} |...| \boldsymbol c_{i_{m}}).
\]

\subsection{Residual Quantization}

Residual quantization (RQ) has a different assumption in that it does not assume the features are independent, but assumes the quantization residuals of the first level quantizer can be further quantized. Thus, it is a stacked quantization model. For the first level, RQ simply uses a $k$-means clustering to quantize the data, and assign them to $k$ centers. For each data point, by subtracting from the assigned centroid, we can collect their residual vectors as $\boldsymbol r_{j} = \boldsymbol x_{j } - q_{1}(\boldsymbol x_{j})$. Then $R = \{\boldsymbol r_{j}, j = 1,...,N\}$ will be used as the input to the next level, and we again use $k$-means\footnote{Although other quantizers can be used, such as a product quantizer, using $k$-means is a basic choice.} to quantize the residual vectors. By repeating this for $h$ times, we can have a sequential product codebook $C = C_{1} \times C_{2} \ldots \times C_{h}$. Given an input vector $\boldsymbol x_{j}$, by computing the nearest center at each level, we can get a sequence of indexes $(i_{1}, i_{2}, \ldots, i_{h})$. The globally nearest center for $\boldsymbol x_{j}$ here becomes a direct sum of all the sub-centers $q(\boldsymbol x_i) = \sum_{l=1}^{h} \boldsymbol c_l$.

Simply considering one level, we represent $\boldsymbol x_{j}$ at level $l$ as $\boldsymbol x_{j}^{l}$ where $\boldsymbol x_{j} = \boldsymbol x_{j}^{1}$. Then the residual vector from level $l$ is 
\begin{equation}
\label{eq:rvec}
\boldsymbol x_{j}^{l+1} = \boldsymbol x_{j}^{l} - q_{l}(\boldsymbol x_{j}^{l}), 
\end{equation}
the MSE for a $h$-level RQ is
\begin{align}
\label{eq:rq_mse}
\text{MSE}(q) = \text{MSE}(q_h) = 1/N \sum_{j=1}^{N} \| \boldsymbol x_{j}^{h} - q_{h}(\boldsymbol x_{j}^{h}) \|_{2}^{2} \nonumber \\
= 1/N \sum_{j=1}^{N} \| \boldsymbol x_j^{h+1} \|_{2}^{2}.
\end{align}
The most important advantage of RQ is that it does not make the unrealistic assumption that the features are statistically independent. In addition, it holds a non-increasing property, that is, adding a level will always reduce the MSE error.

\section{Transformed Residual Quantization}
\label{sectM}

As discussed above, residual quantization recursively performs quantization on residual vectors from previous levels. In other words, all the residual vectors from different clusters are collected and fed into the same quantizer in the next level. As shown in Figure \ref{figure:ORQ-example} (b), the residual clusters can have different shapes, orientations, or scales, which makes effective quantization of them very hard in general. To address this problem, we propose to learn one rotation per cluster to transform the residual clusters, to make them more aligned to each other, and potentially can make more fruitful residual-level quantizers (see Figure \ref{figure:ORQ-example}(c)).

\subsection{Residual Quantization with Cluster-Wise Transforms}

For simplicity, we focus our analysis on a two-level RQ model. We fix the first level to be an $k$-means clustering with $k_{1}$ clusters. And at this point, we are not going to specify the second-level quantizer $q_{2}$ considering our proposed method is general to any quantizer performed in the residual space. In this way, we have a first-level codebook $C^{1}$ with $k_{1}$ cluster centers $\boldsymbol c_{i}^{1}, i=1,...,k_{1}$. We suppose that a vector $\boldsymbol x_{j}$ is assigned to $\boldsymbol c_{i}^{1}$, its residual vector is denoted as 
\begin{equation}
\label{eq:rvec2}
\boldsymbol r_{j} = \boldsymbol x_j - \boldsymbol c_{i}^{1}, 
\end{equation}

A residual cluster $V_{i}$ then is a collection of all residual vectors assigned to its center,
\begin{equation}
\label{eq:rcluster}
V_{i} = \{\boldsymbol r_{j} : q_{1}(\boldsymbol x_{j}) = \boldsymbol c_{i}^{1}\}. 
\end{equation}

The residual set $R = \{\boldsymbol r_{j}: j=1,...,N\}$ forms the input for the next level quantizer $q_{2}$. Finally, the objective function for such a RQ model simply follows either Equation~\ref{eq:rq_mse} or Equation~\ref{eq:pq_mse}, depending on whether another $k$-means or PQ is selected to be the residual quantizer. To alleviate the misalignment of these clusters, we propose to apply cluster-specified transforms right after the first level so that all clusters are aligned for better adaption to the shared residual quantizer. We associate each cluster center $\boldsymbol c_{i}^{1}$ with a transform $T_{i}$, whose inverse $(T_{i})^{-1}$ must exist. Accordingly, a transformed residual cluster is then represented as
\begin{equation}
\label{eq:trans_rcluster}
V_{i}' = \{T_{i}\boldsymbol r_{j} : q_{1}(\boldsymbol x_{j}) = \boldsymbol c_{i}^{1}\}.
\end{equation}

We dubbed the generalized RQ model as transformed residual quantization (TRQ). While our generalization only requires transforms to be invertible, the property of non-increasing MSE error remain valid if transforms are orthogonal matrices, i.e., $(T_i)^T \times T_i = I$, (where $I$ is the identity matrix) because applying orthogonal matrices does not change the MSE error. Yet, in this case, scaling inhomogeneity among residual clusters can not be accounted for by the transforms. Finally, denoting a reproduction cluster as $\widetilde{V}_{i}' = q_{2}(V_{i}')$, and Frobenius norm as $\| . \|_{F}$, the MSE for our TRQ model is given by
\begin{equation}
\label{eq:orq_mse}
\text{MSE} = 1/N \sum_{i=1}^{k_{1}} \| T_{i}V_{i} - \widetilde{V}_{i}' \|_{F},
\end{equation}

\begin{table}[h]
\caption{Complexity comparison with different VQ models, where $m$ is the number of subspaces in PQ model,  $h$ is the number of levels in RQ and TRQ models, and $l$ is the number of levels having transformations involved in our TRQ model; $O_{dist}$ and $O_{tran}$ are operation costs on distance computations and vector transformations respectively.}
\label{table:complexity}
\vskip 0.15in
\begin{center}
\begin{small}
\begin{sc}
\begin{tabular}{lccr}
\hline
%\abovespace\belowspace
Model & \# Cells & Assignment & Memory \\
\hline
%\abovespace
VQ  & $k$ & $kO_{dist}$ & $kD$ \\ 
PQ  & $(k^{\ast})^{m}$ & $k^{\ast}O_{dist}$  & $k^{\ast}D$ \\
RQ  & $(k^{\ast})^{h}$ & $hk^{\ast}O_{dist}$ & $hk^{\ast}D$  \\
%\belowspace
TRQ & $(k^{\ast})^{h}$ & $hk^{\ast}O_{dist}+lO_{tran}$ & $hk^{\ast}D+lk^{\ast}D^{2}$ \\ 
\hline
\end{tabular}
\end{sc}
\end{small}
\end{center}
\vskip -0.1in
\end{table}

The complexity analysis of TRQ is presented in Table \ref{table:complexity} along with the comparison with aforementioned VQ families. TRQ bring extra computational costs with transforms introduced in the model. In section \ref{sectE}, however, we will justify the advantage of TRQ in nearest neighbor search context with consideration of the substantial increase of search accuracy and reasonable control of using additional computational resources.

\subsection{Iterative Residual Cluster Alignment}

The remaining question is how to estimate $T_{i}$'s that would be effective. An intuitive way is to align all clusters to a specific target set of points. The cluster $V_{i}$ with the greatest population, for instance, is widely used in Procrustes shape analysis~\cite{mardia-shape-book}. In this way, a one-to-one matching is required between the two sets, some extra computations then are performed due to the population difference. At the same time, you may have found that a one-to-one matching already exists between $\boldsymbol r_{j}$ and its reproduction $\boldsymbol{\widetilde{r}}_{j} = q_{2}(\boldsymbol r_{j})$. Therefore, a more reasonable choice is a direct alignment from a cluster $V_{i}$ to the reproduction set $\widetilde{V}_{i}$. A byproduct of this alignment is that it directly meets the ultimate goal to minimize the objective function in Equation~\ref{eq:orq_mse}.

Based on these observations, we propose an iterative alignment (IA) approach to finding $T_{i}$'s. Where we iteratively update all the residual sub-codebooks in $C^{2}$ and the orthogonal matrices $T_{i}$'s until MSE converges, or a specified number of iterations has reached. We formulate the two steps as follows,

\textit{Fix $T_{i}$'s, update $C^{2}$}: When $T_{i}$'s are fixed, the procedure is the same as a normal iteration in a vanilla residual quantizer. There are two step involved: 
\begin{enumerate}
\item updating the assignment $I_{j}^{2}$ for each $\boldsymbol r_{j}$, i.e. the indexes of centers $q_{2}(\boldsymbol r_{j})$; 
\item updating the codebook $C^{2}$ according to the new assignment $B = \{ I_{j}^{2} : j=1,...,N\}$.
\end{enumerate}

\textit{Fix $C^{2}$, update $T_{i}$'s}: When the residual codebook $C^{2}$ is fixed, we need to find an optimal orthogonal matrix $T_{i}$ for each residual cluster that
\begin{equation}
\label{eq:alignment_matrix}
T_{i} = \argmin_{\Omega} \| \Omega V_{i} - \widetilde{V}_{i}' \|_{F},
\end{equation}
where $\Omega^{T}\Omega = I$. The equation is solved by the Orthogonal Procrust analysis. The solution is straightforward. Given the covariance matrix $M = V_{i}\widetilde{V}_{i}'$, we use the singular value decomposition (SVD) $M=U\Sigma~V^{\ast}$, to have the objective matrix $T_{i} = UV^{\ast}$.

\subsection{Optimization to Inverted-Index Based ANN Search}
\label{sectM-ANN}

The merits of RQ and PQ techniques for large-scale nearest neighbor search are two folds. For indexing, they produce very large number of landmark points that partition the space, and can be used with efficiency as hashing for fast indexing~\cite{Nister2006, Muja2009, chen2010, YuanL12, YuanL2013, BabenkoL2012}; On the other hand, they can produce a very compact representation of the vectors, and enable us to store huge amount of data in memory ~\cite{Jegou2010, Jegou2011, GeHKS2013, NorouziF2013}. Here we briefly introduce two state of the art ANN search systems, and then discuss optimization to the systems with our IA method. More information about parameter settings and experimental results are given in Section \ref{sectE}.

In the system introduced by Jegou et al.~\cite{Jegou2010, Jegou2011}, the first level is an indexer, which employs the $k$-means to construct an inverted file (IVF). In the second level, a product quantizer is used to compress the residuals to product codes. An approximate search then consists of a shortlist retrieval from the indexer IVF, and candidates re-ranking by the product codes. As database size increases, however, the inverted file becomes a bottleneck since the complexity for a $k$-means codebook learning is $O(Nk)$ and a vector assignment is $O(k)$. Where $k$ is the codebook size or the number of cells in IVF. In order to address this issue, Babenko et al. \cite{BabenkoL2012} introduced the inverted multi-index to replace the IVF in above system. The new indexer employs a novel algorithm to jointly sort product cells in a PQ model. The complexities are reduced to $O(Nk^{\frac{1}{m}})$ for PQ codebook learning and $O(k^{\frac{1}{m}})$ for an vector assignment. It has been shown that both search speed and quality were improved. Ge et al. \cite{GeHKS2013} recently applied their optimized PQ (OPQ) approach to the later system. The two levels are both optimized because they are essentially two PQ models.

Both of the systems are indeed based on hybrid models of residual product quantization. In the IVF system, the first level is a flat $k$-means, and the residual level is an ordinary product quantizer. At the same time, the inverted multi-index seems to be more complicated since two-level PQ models involved. However, we could simply see it as $m_1$ residual product quantizers independently produced from the $m_1$ subspaces in the multi-index structure. To optimize such residual product quantizers by our IA method, we only need to pay attention to the steps of the assignment and residual codebook $C^{2}$ updating, to make them follow the way of an ordinary PQ codebook learning.

\section{Experiments}
\label{sectE}

In this section, we examine the proposed TRQ model's performance on several large scale benchmark datasets. We consider both quantization distortion (i.e. mean squared error (MSE)) and ANN search. The first task is a direct measurement of the quantization error of different models, and the second task is a real world ANN search application of these quantization models.

\begin{table}[!ht]
\caption{Statistics of datasets used.}
\label{table:datasets}
\vskip 0.15in
\begin{center}
\begin{small}
\begin{sc}
\begin{tabular}{lcccr}
\hline
%\abovespace\belowspace
Data set & \# Dim. & \# Base & \# Training & \# Query \\ 
\hline
%\abovespace
MNIST   & 784 & 60K & 60K & 10K\\ 
SIFT1M    & 128 & $10^{6}$ & 100K & 10K\\ 
GIST1M   & 960 & $10^{6}$ & 500K & 1K\\ 
%\belowspace
SIFT1B   & 128 & $10^{9}$ & $10^{8}$ & 10K \\ 
\hline
\end{tabular}
\end{sc}
\end{small}
\end{center}
\vskip -0.1in
\end{table}

\begin{figure*}[!ht]
\begin{center}
 \subfloat{\label{fig:MSE-mnist60k}
 \includegraphics[trim=2.1cm 16.3cm 2.2cm 6.9cm,clip=true, width=.95\linewidth] 
 {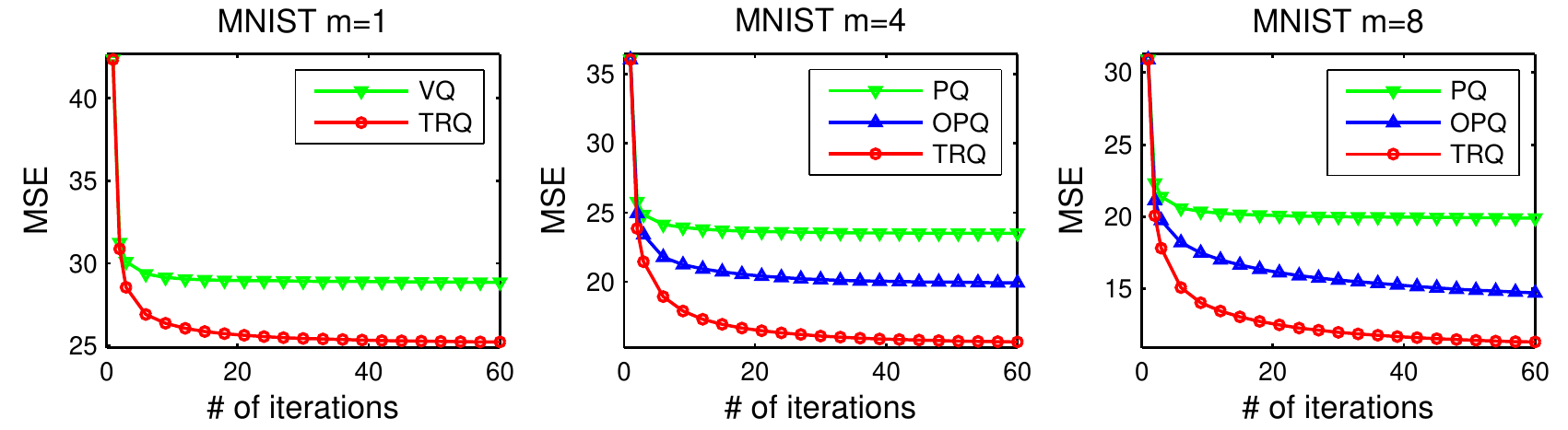}} \\
 \subfloat{\label{fig:MSE-sift1m}
 \includegraphics[trim=2.1cm 16.3cm 2.2cm 6.9cm,clip=true, width=.95\textwidth]
 {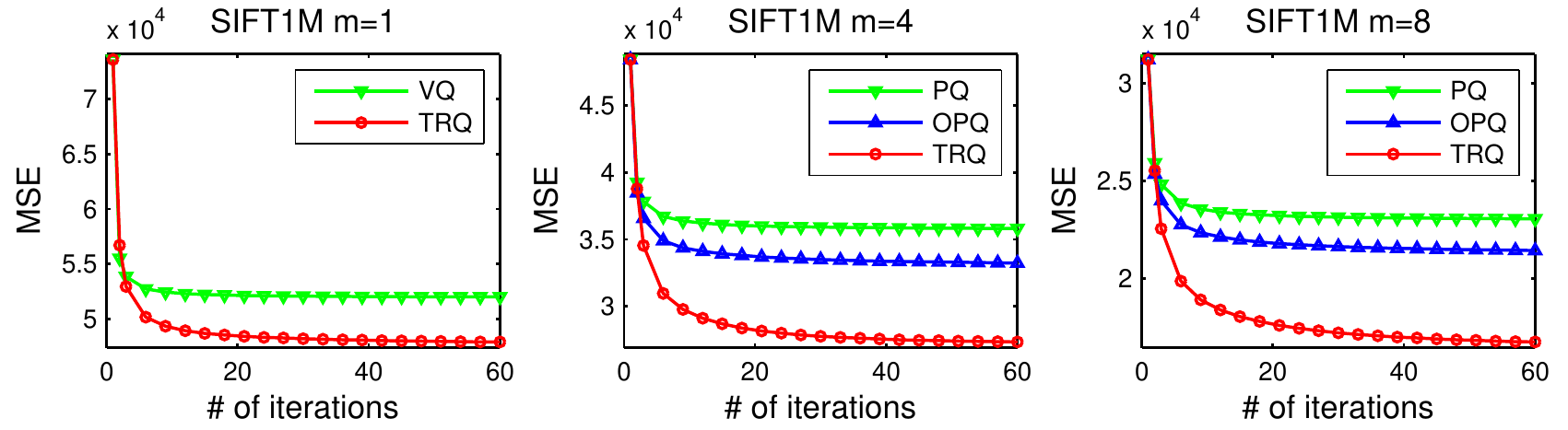}} \\
 \subfloat{\label{fig:MSE-gist1m}
 \includegraphics[trim=2.1cm 16.3cm 2.2cm 6.9cm,clip=true, width=.95\textwidth]
 {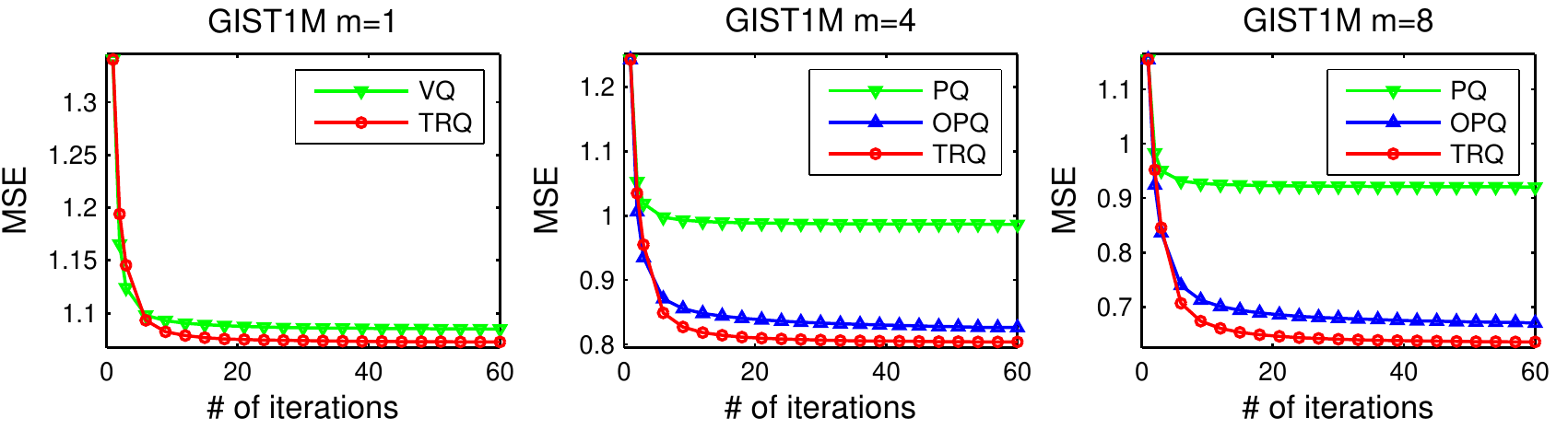}}
\end{center}
\caption{Comparative evaluation about distortion performance. The first level quantizers are fixed across all models so we only examine the behaviors on the residual quantizers. For SIFT1M and GIST1M, we use $k1 = 32$ and $k2 = 256$ that are numbers of clusters used in two levels. While for MNIST, $k1 = 10$ and $k2 = 64$. And $m$-- is the number of feature groups for residual quantizers. In addition, the codebook trainings are directly performed on the base sets.}
\label{fig:MSE}
\end{figure*}

We have used four large scale datasets in our evaluation. The data types include images of hand-written digits from MNIST, local image descriptors SFIT~\cite{Lowe2004}, and global image descriptors GIST~\cite{Oliva2001}. See Table~\ref{table:datasets} for statistics of the datasets. We will present in-depth experimental analysis on three relatively small datasets (MNIST~\cite{LeCunCB1998}, GIST1M, SIFT1M~\cite{Jegou2010}), and report large-scale ANN results on the largest SIFT1B dataset~\cite{Jegou2011}.

In our experiments, we mainly consider two strong quantization baselines. The first method is product quantization (PQ) \cite{Jegou2010} which groups features into different groups, and performs quantization separately. The second method is optimized product quantization (OPQ) \cite{NorouziF2013}, which learns a rotation matrix for PQ to make feature groups more independent with each other, and to make quantization error smaller. This method has been shown to produce state of the art performance for ANN search.

\subsection{Evaluation on Quantization Distortion}

We first report quantization distortion for each model on different datasets, and we use mean squared error (MSE) as the distortion measurement. We only consider a 2 level quantization model in this section (no additional stacked layers for RQ and TRQ), which will help us better evaluate the quantization distortion. The parameter settings are given in Figure~\ref{fig:MSE}. For PQ, when the number of feature groups $m=1$, it does not perform any grouping to the feature dimensions, and is reduced to an ordinary VQ ($k$-means clustering).

Figure~\ref{fig:MSE} shows clearly that our proposed method TRQ has achieved a very significant gain over all other methods. Both TRQ and OPQ work better than the vanilla PQ model, which shows learning these optimal rotations has a clear advantage. Our method further improves OPQ by a significant margin. This is probably because our model does not make any independence assumptions about the feature dimensions or feature groups, while PQ and OPQ heavily rely on this assumption, which is usually too strong for real-world data. It also shows
 that the gain of our methods is much more significant on MNIST and SIFT than that on GIST. This is probably because the feature dimensions of GIST are more independent, as has been shown and verified in \cite{Jegou2010}. In general, on all three datasets, our proposed TRQ method has achieved consistent and substantial gain over all other state of the art methods.

\subsection{Evaluation on ANN Search}

Our second experiment applies our method to optimize the ANN search system from~\cite{Jegou2010}. The system is a typical two-level RQ model described in Section~\ref{sectM}, which first applies $k$-means clustering and cluster the dataset into $k$ clusters. All the residual vectors are then passed to a PQ to learn the compact codes. During a search, for a query point $\boldsymbol y$, we first compute its distance to all the first-level cluster centers, and probe $\boldsymbol y$ into the nearest $w$ clusters, and then compute corresponding residuals $\boldsymbol r_{j}, j=1,...,w$ for $\boldsymbol y$. If $\boldsymbol r_{j}$ is from cluster $i$, then we need to rotate it as $\boldsymbol r_{j}' = T_{i}\boldsymbol r_{j}$ in our TRQ model. Finally, each $\boldsymbol r_{j}'$ is used to re-rank all candidates in its cluster through their PQ-codes. Depending on the dataset size, we use $w=6$ for SIFT1M and GIST1M, and $w=2$ for MNIST. We use Recall@R to measure the search quality, which represents the recall of the nearest neighbor within the top R candidates. 

\begin{figure*}[h]
\begin{center}
 \includegraphics[trim=2.1cm 16.4cm 2.2cm 6.9cm,clip=true, width=.95\textwidth]{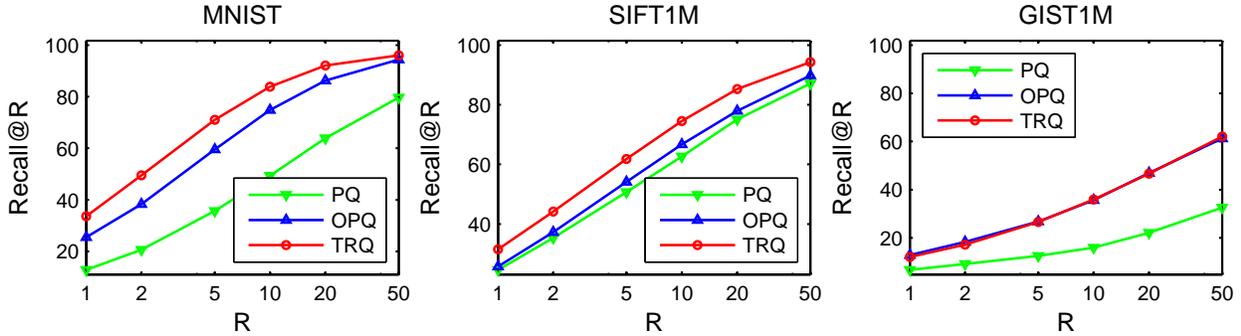}
\end{center}
\caption{ANN search evaluation on different datasets. The parameters used are exactly the same as in Fig. \ref{fig:MSE}, except that here $m=8$ is fixed for all.}
\label{fig:ann-small}
\end{figure*}

Figure \ref{fig:ann-small} shows a similar trend as that of the previous section. Our TRQ method has achieved significantly better performance than the other two methods. For instance, on SIFT1M, we have achieved 31.51\% on Recall@1, while OPQ achieved 25.77\% and PQ 24.61\%, respectively. Similar to the quantization distortion experiments, the gain of our method is much smaller on the GIST1M. This is probably because the feature dimensions of GIST are more independent (as verified by \cite{Jegou2010}). The true underlying reasons
are to be further investigated.

\subsection{Large-Scale Experiments with The Inverted Multi-Index}

The last experiment shown in this paper is to integrate our TRQ model with the inverted multi-index~\cite{BabenkoL2012} to improve its search accuracy on SIFT1B dataset. From Table \ref{table:sift1b-result}, we can see that the search accuracy was substantially improved by our TRQ , which is up to $8\%$ Recall@1 comparing to OPQ. 

As the discussion in Section \ref{sectM-ANN}, the basic search steps are similar to the above IVF system~\cite{Jegou2010}, which include shortlist retrieval and candidates re-ranking by PQ-codes. We here refer interested readers to~\cite{BabenkoL2012} and~\cite{GeHKS2013} for more details about the inverted multi-index and its optimized version called OMulti~\cite{GeHKS2013}. Here we are going to discuss more details about the search procedure when our TRQ integrated with OMulti, and more comparative results with other state of the art methods.

\begin{table}[!ht]
\caption{ANN search comparisons between our model TRQ with OPQ \cite{GeHKS2014} on SIFT1B with 16-byte codes ($m$=16) per vector. This table corresponds to Table 5 in \cite{GeHKS2014}. $T$ is the length of shortlist retrieved by the optimized multi-indexer OMulti \cite{BabenkoL2012, GeHKS2014} for final re-ranking. The time is average search time per query that consists of shortlist retrieval and re-ranking.}
\label{table:sift1b-result}
\vskip 0.15in
\begin{center}
\begin{small}
\begin{sc}
\begin{tabular}{lccccr}
\hline
%\abovespace\belowspace
Method & $T$ & R@1 & R@10 & R@50 & $t$(ms) \\
\hline
%\abovespace
 OPQ & 10K & 0.359 & 0.734 & 0.791 & 4.2 \\
 TRQ & 10K & 0.426 & 0.769 & 0.795 & 4.8 \\ 
 OPQ & 30K & 0.379 & 0.818 & 0.907 & 7.9 \\
 TRQ & 30K & 0.446 & 0.868 & 0.914 & 8.6 \\ 
 OPQ & 100K & 0.385 & 0.851 & 0.961 & 18.2 \\
%\belowspace
 TRQ & 100K & 0.465 & 0.911 & 0.972 & 19.8 \\ 
\hline
\end{tabular}
\end{sc}
\end{small}
\end{center}
\vskip -0.1in
\end{table}

\textbf{Parameters} There are mainly four parameters involved, where $m_{1}$ and $m_{2}$ are used to represent the numbers of feature groups for the two different level PQ models, while $k_{1}$ and $k_{2}$ as the numbers of centers. We choose $m_{1}=2$ and $m_{2}=16$, with $k_{1}=16384$ and $k_{2}=256$, to be consistent with~\cite{BabenkoL2012} and~\cite{GeHKS2013}. Since the multi-index has split the features to two groups (see $m_1=2$), the system can simply be seen as a concatenation of two independent residual product quantization models, and each of them has $m_{1}=1$ and $m_{2}=8$ while $k_1$ and $k_2$ are retained. In our case, they are optimized as two TRQ models. 

From now on, we only need to present how one of the TRQ models works during a search. Note that, we are integrating TRQ with OMulti, so the whole raw space has been transformed by, say $T^{1}$, the transformation learnt in OMulti. A query $\boldsymbol y$ becomes $\boldsymbol y'=T^{1} \boldsymbol y$ before it probes to the first-level clusters. We split $y'$ equally to two segments, $\boldsymbol y'= \boldsymbol y_{1}' | \boldsymbol y_{2}'$. Here we show only what happens to $\boldsymbol y_{1}'$; it then will be probed into the left subspace. If it is assigned to the sub-center $\boldsymbol c_{1i}^{2}$, in our setting, the residual for $\boldsymbol y_{1}'$ is $\boldsymbol r_{1}'' = T_{1i}^{2} \boldsymbol r_{1}' = T_{1i}^{2}(\boldsymbol y_{1}' - \boldsymbol c_{1i}^{2})$. Finally, the concatenation $\boldsymbol r_{1}'' | \boldsymbol r_{2}''$ are used for re-ranking PQ-codes on the shortlist that is provided by the indexer OMulti. 

\textbf{Computation-Wise Discussion} A potential concern might be centered on the extra computational and storage costs required by the proposed TRQ model. TRQ applies transformations in the residual spaces and aims to improve the quantization quality and furthermore the search accuracy in the ANN problem. It slightly increases memory usage due to the local transformations. In the setting for Table \ref{table:sift1b-result}, for instance, it requires about 2.5\% extra memory space for all the transformation matrices relative to PQ or OPQ model. This is acceptable with the substantially increased search accuracy considered. 

Even though our presented results have been restricted to comparisons with PQ and OPQ, we here give a short discussion about comparison between our method with LOPQ~\cite{KalantidisA2014} considering the common feature of the locally learned transformations. LOPQ learn different PQ codebooks for each cluster while our TRQ share just one PQ codebook among all clusters. This results in LOPQ use much more extra memory than TRQ dose. In the same setting as Table \ref{table:sift1b-result} LOPQ produce 16384 PQ codebooks that takes 2 GB (about 10\%) extra memory compared to ours and other methods. Taking advantage of the considerable memory usage, LOPQ have achieved about 2\% higher recall than our TRQ on this SIFT1B dataset.

Computationally, the number of transformations or matrix multiplications in both TRQ and LOPQ for each query is determined by the number of visited cells from the first level index; it is usually less than 100 out of 16384 (in Table 2). The results in Table \ref{table:sift1b-result} show that the extra computational costs can be well justified given the significant improvements on search accuracy. We also use only 10 millions of the training points to speed up the codebooks learning in both of OPQ and TRQ. All the results have been produced using a single thread with an Intel(R) Xeon(R) CPU @ 3.40GHz with 25.6M cache size, and 128 GB RAM. All methods have been implemented in MATLAB and C (for codebook learning and database encoding) and C++ (for searching).

\section{Conclusions and Future Directions}

Product quantization and residual quantization are the well-known quantization models that reduce both computation and storage complexities. We have seen that different ways to optimize these models are becoming active and fruitful in the computer vision area. In this paper, we have presented a novel approach to reducing quantization error of residual quantization for fast nearest neighbor search. We have also presented two hybrid searching architectures that are based on product quantization and the proposed transformed residual quantization. Our experiments show that the proposed TRQ has achieved significantly smaller quantization error than previous methods, which clearly demonstrates the benefits of the proposed model. Also, in large-scale ANN experiments on 1 billion vectors, our method has achieved significantly better accuracy than previous methods.

Despite that we have restricted our attention to the ANN search problem in this paper, the approach can be extensively used for many other problems that require efficient and effective very large-scale clustering. For examples, $k$NN graph construction, and feature matching for image retrieval, and for object recognition to name a few. In addition, how to integrate our transformation-based approach with the global codebook learning approaches arises to be a very interesting question for residual quantization optimization. These become our future research directions.

\end{document}